\documentclass[pdflatex,sn-mathphys-num]{sn-jnl}

\usepackage{graphicx}%
\usepackage{multirow}%
\usepackage{amsmath,amssymb,amsfonts}%
\usepackage{amsthm}%
\usepackage{mathrsfs}%
\usepackage[title]{appendix}%
\usepackage{xcolor}%
\usepackage{textcomp}%
\usepackage{manyfoot}%
\usepackage{booktabs}%
\usepackage{algorithm}%
\usepackage{algorithmicx}%
\usepackage{algpseudocode}%
\usepackage{listings}%

\theoremstyle{thmstyleone}%
\theoremstyle{thmstyletwo}%

\theoremstyle{thmstylethree}%

\begin{document}

\title[What is the goal of unsupervised learning?]{What is the goal of unsupervised machine learning?}


\author*{\fnm{Aapo} \sur{Hyvärinen}}\email{aapo.hyvarinen@helsinki.fi}

\affil{\orgdiv{Department of Computer Science}, \orgname{University of Helsinki},\\ \orgaddress{\street{Pietari Kalmin katu 5}, \city{Helsinki}, \postcode{00560}, \country{Finland}}}


\abstract{Unsupervised learning is one of the main branches of machine learning. Here I argue that unlike the other branches of machine learning (supervised and reinforcement learning), unsupervised learning is a rather heterogenous field that can serve several different goals. It seems futile to try to define one single goal for unsupervised learning. I identify four different goals for unsupervised learning: 1) \pdftext, 2) \samplingtext, 3) \featurestext, and  4) \understandingtext.}

\keywords{Unsupervised learning, Deep learning, Probabilistic models}



\newcommand{\ncov}{\boldsymbol{\Sigma}_\n}
\newcommand{\scov}{\boldsymbol{\Sigma}_\s}
\newcommand{\xcov}{\boldsymbol{\Sigma}_\x}
\newcommand{\nablas}{\nabla_\s}
\newcommand{\nablax}{\nabla_\x}
\newcommand{\Jx}{\mathbf{J}_\x}
\newcommand{\ns}{\sigma^2}
\newcommand{\px}{p_\x}
\newcommand{\eps}{\epsilon}
\newcommand{\ps}{p_\s}
\newcommand{\py}{p_\y}
\newcommand{\xscore}{\boldsymbol{\Psi}_\x}
\newcommand{\manifold}{\mathcal{M}}
\newcommand{\hex}{\mathbf{H}_\x}
\newcommand{\sscore}{\boldsymbol{\Psi}_\s}
\newcommand{\yscore}{\boldsymbol{\Psi}_\y}
\newcommand{\W}{{\bf W}}
\newcommand{\V}{{\bf V}}
\renewcommand{\H}{{\bf H}}
\newcommand{\C}{{\bf C}}
\newcommand{\I}{{\bf I}}
\newcommand{\tr}{\text{tr}}
\newcommand{\Ctwo}{\mathcal{C}^2}
\newcommand{\A}{{\bf A}}
\newcommand{\R}{\mathbb{R}}
\newcommand{\X}{{\bf X}}
\newcommand{\M}{{\bf M}}
\newcommand{\Tb}{{\bf T}}
\newcommand{\Tbar}{\bar{\bf T}}
\newcommand{\x}{{\bf x}}
\newcommand{\f}{{\bf f}} 
\newcommand{\h}{{\bf h}}
\newcommand{\g}{{\bf g}}
\newcommand{\ginv}{{\bf g}_+}
\newcommand{\J}{{\bf J}}
\newcommand{\U}{{\bf U}}
\newcommand{\xs}{{\bf \tilde{x}}}
\newcommand{\z}{{\bf z}}
\newcommand{\y}{{\bf y}}
\newcommand{\df}{\text{d}}
\newcommand{\E}{\mathrm{E}}
\newcommand{\Ehat}{\hat{E}}
\newcommand{\at}{{\bf \tilde{a}}}
\newcommand{\bt}{{\bf \tilde{b}}}
\newcommand{\wt}{{\bf \tilde{w}}}
\newcommand{\vt}{{\bf \tilde{v}}}
\newcommand{\Mt}{{\bf \tilde{M}}}
\newcommand{\Dt}{{\bf \tilde{D}}}
\newcommand{\Ct}{{\bf \tilde{C}}}
\newcommand{\w}{{\bf w}}
\newcommand{\n}{{\bf n}}
\renewcommand{\a}{{\bf a}}
\renewcommand{\b}{{\bf b}}
\renewcommand{\v}{{\bf v}}
\newcommand{\Ss}{{\bf S}}
\newcommand{\mm}{{\bf m}}
\newcommand{\s}{{\bf s}}
\newcommand{\K}{{\bf K}}
\newcommand{\D}{{\bf D}}
\newcommand{\B}{{\bf B}}
\newcommand{\bb}{{\bf b}}
\newcommand{\cc}{{\bf c}}
\newcommand{\qq}{{\bf q}}
\newcommand{\kurt}{\text{kurt}}
\newcommand{\mub}{{\boldsymbol{\mu}}}
\newcommand{\psib}{{\boldsymbol{\psi}}}
\newcommand{\sigmab}{{\boldsymbol{\sigma}}}
\newcommand{\alphab}{{\boldsymbol{\alpha}}}
\newcommand{\uu}{{\bf u}}
\newcommand{\etab}{{\boldsymbol{\eta}}}
\newcommand{\nub}{{\boldsymbol{\nu}}}
\newcommand{\thetab}{{\boldsymbol{\theta}}}
\newcommand{\thetabhat}{{\boldsymbol{\hat{\theta}}}}
\newcommand{\etabhat}{{\boldsymbol{\hat{\eta}}}}
\newcommand{\xib}{{\boldsymbol{\xi}}}
\newcommand{\pro}{\bar{\x}}
\newcommand{\Gram}{\boldsymbol{\Phi}}
\newcommand{\dist}{\boldsymbol{\Psi}}
\newcommand{\env}{\text{env}}
\newcommand{\half}{\frac{1}{2}}
\newcommand{\imxw}{\hat{\z}^{\text{Im}}}
\newcommand{\rexw}{\hat{\z}^{\text{Re}}}
\newcommand{\imxi}{\hat{\x}^{\text{Im}}_i}
\newcommand{\rexi}{\hat{\x}^{\text{Re}}_i}
\newcommand{\exclude}[1]{}
\newcommand{\ft}[1]{\hat{#1}}
\newcommand{\xalt}{{\bf \tilde{x}}}
\newcommand{\yalt}{{\bf \tilde{y}}}
\newcommand{\zalt}{{\bf \tilde{z}}}
\newcommand{\pdf}{\#1}
\newcommand{\sampling}{\#2}
\newcommand{\features}{\#3}
\newcommand{\understanding}{\#4}
\newcommand{\pdftext}{Estimating the distribution}
\newcommand{\samplingtext}{Generating new data points}
\newcommand{\featurestext}{Extracting features for downstream tasks}
\newcommand{\understandingtext}{Understanding the data}

\newcommand{\textnew}[1]{\color{black} #1 \color{black}}

\maketitle

\section{Introduction}

Machine learning is classically divided into supervised learning, unsupervised learning, and reinforcement learning. Roughly speaking, in supervised learning, some kind of prediction accuracy is maximized, and in reinforcement learning, the total reward is maximized. In those cases, we have relatively well-defined goals on which there is a large consensus.

But what is maximized in unsupervised learning, or what is its goal? The starting point in this paper is that the answer to that question is complicated, even controversial. It does not seem possible to point out one single goal even on a conceptual level, let alone a single objective function that would be maximized.

Instead, my claim here is that unsupervised learning has several different goals which are quite different from each other. Some of them are orthogonal in the sense that trying to reach one goal does not in any way move towards another goal. In fact, the goals may even be a bit contradictory with each other.

\textnew{Note that the discussion in this paper also applies to self-supervised learning (SSL). While there are different viewpoints in the literature, I consider SSL as an algorithmic technique that uses supervised algorithms to accomplish the goals of unsupervised learning. Thus, the goals themselves are the same for SSL and unsupervised learning.}

I tentatively specify four different goals for unsupervised learning. There is no claim that this list is complete, and other goals could probably be identified. My list includes the following goals:
\begin{enumerate}
\item \pdftext
\item \samplingtext
\item \featurestext
\item \understandingtext
\end{enumerate}
I will try to justify why these can be seen as separate goals in unsupervised learning, and how they relate to each other. Some of the basic methods for each goal are pointed out, but this is not an attempt to make a comprehensive review of the topic. Particular emphasis is here laid on how such learning methods could be validated in view of each of the goals. Related analysis has been presented by \cite{theis2016note}.

\paragraph{Definitions}

Assume we have an observed data set $\X=\{\x_1,\ldots,\x_N\}$, where each data point $\x_i$ is in a real space $\R^n$. In the case of supervised learning, it is further assumed that to each $\x_i$ is associated a label or target $\y_i$ but this is not available in the basic unsupervised case.
We further take a probabilistic viewpoint, and make some further assumptions: a) the $\x_i$ are observations of a random variable $\x$ that follows the pdf $p_\x$; b) the observations are independent of each other, in other words $\x_i$ form an i.i.d. sample. The latter assumption could be relaxed, but it presents the most fundamental case.

\section{Proposed list of goals for unsupervised learning}

\subsection*{Goal \pdf: \pdftext}

\paragraph{Problem statement}

Perhaps the most fundamental goal of unsupervised learning is learning the probability distribution of the data. This typically takes of the form of learning the probability density function $p_\x$, or its logarithm, from the data set $\X$. Alternatively, in some recent developments, the score function $\nabla_\x \log p_\x$ is the target of learning \cite{Hyva05JMLR,vincent2011connection,song2019generative}. 

From an extreme probabilistic viewpoint, that is all that is ever needed. In theory, even supervised learning can be solved if the relevant densities are known. For example, consider classification with two classes, with pdf's $p_1$ and $p_2$. After learning the densities separately for the two classes, the classification problem can be solved by assigning any data point $\x$ to class 1 if $p_1(\x)>p_2(\x)$ and to class 4 otherwise. This is a special case of using the density ratio $p_1/p_2$ which has many other utilities as well \cite{sugiyama2012density}. In fact, even basic logistic regression converges to compute exactly this ratio \cite{Gutmann12JMLR}.

Furthermore, a more general form of the regression problem can be formulated by considering the conditional pdf $p(\y|\x)$. This can be accomplished, for example, by learning the pdf of a data vector where the regressors and regressands are concatenated: $\z_i=(\x_i,\y_i)$. Now for any new \textnew{(conditioning)} data point $\x_k$, we only need to consider the function $g(\etab)=p_\z(\x_k,\etab)$ which is equal to the conditional pdf $p(\y|\x)$ up to a constant (which is $1/p_\x(\x)$). \textnew{Below, it will be seen how such conditioning is also useful in data generation.}
 
The downside with merely performing density estimation is that it may not enable us to understand the structure of the data, since the density approximator is often an uninterpretable black box given by a neural network \cite{vincent2011connection,song2021train}. Nor is it clear if it is a practical method for solving any supervised tasks; the connections shown above are of a more theoretical nature. These problems will serve as motivations for the other goals below.

\paragraph{Methods}

The classic statistical method for general (``non-parametric'') density estimation is the kernel density estimator \cite{chen2017tutorial}; unfortunately, such methods do not work well in high dimensions. Methods based on reproducing kernel Hilbert spaces are a related option \cite{Bharath17,wenliang19a}.

Deep learning research focuses on Energy-Based Modelling and Score-Based Modelling, using methods stemming from score matching \cite{Hyva05JMLR} or noise-contrastive estimation \cite{Gutmann12JMLR}. Such methods tend to be state-of-the-art \textnew{(or close)} in high-dimensional spaces, although they each pose their specific technical problems, as discussed by \cite{song2021train}.

Normalizing flows \cite{papamakarios2021normalizing} are another approach.
In general, it is very difficult to use using maximum likelihood to estimate the pdf approximated by a a neural network. Normalizing flows make this possible by imposing suitable restrictions on the neural network. 
However, the quality of the density approximation may suffer from such restrictions.

\paragraph{Validation}

Now we discuss the problem of validating a density estimation method after it has been learned, i.e.\ showing that it works, or not; or how it compares to other methods. As in the rest of this paper, the discussion focuses on real-valued vector data.

In some cases, the density estimation methods can be validated by computing the objective function on a held-out test data set. Suppose we have two neural networks defined by different architectures, both trained with one of the estimation methods above (say, score matching) to approximate the log-pdf. For those methods that work by minimizing an objective function, one can compute the objective \textnew{(e.g.\ the score matching objective)} on the test data set, which enables comparison of the two neural networks.

However, the objective function values usually have no clear interpretation and it is therefore difficult to say if the value of the objective is small or large, i.e.\ if the model fit is good or not. The same problem actually occurs with ordinary maximum likelihood: It is usually not possible to say if a given value of likelihood is high or low in some intuitive sense. This is in contrast to classification, where it is clear that, say, 99\% classification accuracy is quite high in most applications. Thus, this validation method can only perform a comparison between two architectures or similar cases. 


Another widely-used validation method is to use the learned distribution to generate new data by sampling, which will be discussed below. However, this is merely redefining the problem as the validation of data generation, which turns out to be very difficult, as will also be discussed below.

\subsection*{Goal \sampling: \samplingtext}

\paragraph{Problem statement}

A rather recent development in machine learning is generative AI, which is typically formalized as sampling a data point from the pdf of the data $p_\x$. In practice, the distribution is conditioned by a prompt or similar input $\uu$, so the sample is actually generated from $p_{\x|\uu}$. For simplicity, though, we talk about sampling from $p_\x$ in what follows.

While such sampling is a classic problem in statistics, modern generative AI has emphasized the case where $p_\x$ is not given as an analytical formula, as in classical theory, but the sampling system is learned from the observed data set $\x_i$ alone. Such a sampling framework is, in particular, the basis for image generation methods. (Text generation uses rather different frameworks and is not considered here.)

\paragraph{Methods}

\textnew{Recent real-space data generation methods in AI} typically use a two-step procedure, where an estimate of the pdf $p_\x$ is first obtained, perhaps in the form of the score function $\nabla_\x \log p_\x$ \cite{Hyva05JMLR,song2021train,vincent2011connection}. Thus, this first step is utilizing the methods described above in relation to Goal~\pdf\ above.
As the second step, any method for sampling from a given pdf (or score function) is used. Typical sampling methods include Langevin methods \cite{song2019generative,Hyva25TMLR} and (generative) diffusion models \cite{croitoru2023diffusion,yang2023diffusion}. Such sampling can also happen in a feature space found by another unsupervised learning system \cite{rombach2022high}.

But sampling can also be performed by special methods that do nothing else than sampling. One of the triggers of the generative AI boom was the method of Generative Adversarial Networks or GANs \cite{Goodfellow14}, which directly learns to generate new data points from $p_\x$  without \textnew{performing any density estimation or similar.} In fact, many latent variables models, such a variational autoencoders and nonlinear independent component analysis \cite{Kingma14,Hyva23patterns} can also generate new data points; however, the results may not be competitive with methods that are specifically created or optimized for data generation.

\paragraph{Validation}

Validation of such data generation is quite challenging. There is no simple obvious measure that would tell how closely a generated data point, or even a set of generated points, follows the distribution of the original data set $\X$. Another caveat is that it is important to make sure the sampling system is not just copying the training data points.

In the case of image generation, visual inspection of the generated images is an obvious choice, but this is clearly not a rigorous, quantitative measure. Researchers have developed some standard measures based on benchmark data sets, so-called ``inception'' scores or distances \cite{salimans2016improved,heusel2017gans}. These typically measure how the generated images fit with the classes of the benchmark dataset such as ImageNet. However, they are restricted to domains where a labelled benchmark dataset is available, such as images.

A more general possibility would be to check if a deep neural network can learn to discriminate the generated data points from the original data points. However, since this measure is in fact optimized in the GAN learning process, using it as a new measure does not give any new information if GAN is used as the generator. Moreover, this measure depends on the architecture and training procedure of the neural network that is trained in calculating the measure. Thus, a general solution to this validation problem is hard to find.

\subsection*{Goal \features: \featurestext}

\paragraph{Problem statement}

Feature extraction means that we learn a function $\f:\R^n\rightarrow\R^m$, called the feature extractor, such that the features in $\z=\f(\x)$ are somehow useful. This is a special case of \textit{representation learning} \cite{bengio2013representation}. The utility of such features can take many forms, but we consider here the particular case where the features are used for further learning in another task, called the ``downstream'' task.
The downstream task is typically supervised, but it could also be reinforcement learning, and it could even be a further unsupervised task, based on the two goals above.
Here we focus on a supervised downstream task, in which case this is closely related to what is called \textit{semi-supervised learning}.

The motivation of semi-supervised learning is that it is often easy to obtain a large number of data points $\x_i$ that have no labels or targets $\y_i$, while any labels or targets are more difficult or expensive to obtain. Thus, we can have a large dataset without labels, and only a small data set that has proper labels. 
For example, one can find, on the Internet, huge amounts of images and text that have no specific information on their contents (e.g.\ does the image depict a cat or or a dog); these form the unlabelled data set $\x_i$. Labels can be created, for example, by human experts but this is expensive and can only be done for a small subset. Let us call the subset with labels $\xalt_j$ with labels $\yalt_j$. 

Now, the goal of unsupervised learning in this case is to first learn to transform the data into features  $\z=\f(\x)$ based on the unlabelled data alone. The features $\z$ should be such that it is much easier to perform the ``downstream'' supervised learning task of predicting $\y$ from $\x$ by using the \textnew{features} $\z$ to predict $\y$ instead of the original $\x$. Thus, learning the downsteam task becomes possible even using the small labelled dataset consisting of $\xalt$ and $\yalt$ and the features $\zalt$. 

\paragraph{Methods}

A large number of methods fall into this category. In fact, almost any neural network trained in an unsupervised manner can be hoped to give useful features in its hidden layers.
Recent research has further produces a large number of methods based on \textit{self-supervised learning (SSL)} principles. Among SSL we can distinguish two main strands: autoencoders and contrastive learning.

\textit{Autoencoders} \cite{bank2023autoencoders} are in fact one of the oldest methods for unsupervised learning, having their roots in the early neural networks research in the 1980's. The basic principle is to predict $\x$ by a neural network that takes the same $\x$ as input, but has a bottleneck such as a hidden layer with a small number of units. Autoencoders have been recently revived in the form of variational autoencoders or VAEs \cite{Kingma14} which provide a principled probabilistic framework. However, it is often not clear why such a probabilistic VAE should be preferred to a very basic autoencoder. Recent research has also emphasized methods where instead of a bottleneck in the neural network, the input is corrupted, for example by adding noise \cite{Vincent2010stacked}. 

\textit{Contrastive learning} is a term used in quite different meanings by different authors. The main thing most of these methods have in common is that they transform the unsupervised learning into a classification problem (as opposed to ordinary regression as in autoencoders). A fundamental probabilistic framework is provided by 
noise-contrastive learning \cite{Gutmann12JMLR}, 
while a rather different approach is provided by SimCLR  \cite{Chen2020simple}.
Contrastive learning approaches have also been used in the context of nonlinear ICA \cite{Hyva23patterns}.

For downstream tasks related to reinforcement learning, the seminal methods include the successor representation \cite{Dayan93} and proto-value functions \cite{Mahadevan08}. \textnew{While these methods may not usually be considered unsupervised learning methods, they explore the action space without referring to the rewards, and are in that sense unsupervised.}

\paragraph{Validation}

Validation of whether this goal has been achieved is straightforward in the framework of semi-supervised learning: We only need to measure the performance on the downstream task by some conventional measure such as classification accuracy. Likewise, comparing several unsupervised feature extraction methods is straightforward using the same measure. Thus, the semi-supervised learning framework is interesting also in the sense that its performance can be easily measured; it provides one objective way of comparing different unsupervised learning algorithms. However, our emphasis here is that this is only one possible goal of unsupervised learning, and it would be ill-advised to evaluate or compare any unsupervised learning methods by this metric.

\subsection*{Goal \understanding: \understandingtext}

\paragraph{Problem statement}

Especially in scientific data analysis, the main goal of machine learning is to further the understanding of the structure of the data and the phenomena being measured.  It is actually quite commonplace to train a classifier on scientific data, where the classes correspond to something like experimental conditions. But even if such a classifier is trained, the real goal may not be to predict anything, but to understand how the prediction is performed by the system. If a linear regressor or a linear classifier is trained on the data, interpretation of the regression coefficients is straightforward; if a neural network is used, understanding how it functions is notoriously difficult. 

Regarding unsupervised learning, a common approach is to find components or factors as a basis for understanding the data. These features may  not be very different from  the features considered in Goal~\features\ in semi-supervised learning. Thus, the same components may sometimes be useful for both the current goal and Goal~\features. In fact, understanding and interpretation by a human investigator could be seen as another downstream task, at least on a very abstract level. But there is actually a clear difference between the goals: the features in Goal~\features\ may not facilitate understanding the data if the feature extraction is performed by a complex neural network whose computation is too difficult to understand by humans; \textnew{ or simply, if the number and size of the features is overwhelmingly large.} Thus, it is justified to consider understanding the data a distinct goal of learning.

\textit{Identifiability} is a key concept with this goal  \cite{Hyva23patterns,Hyva24AISM}. It means that the features to be learned are unique and well-defined. Many widely used feature extraction methods only give the features up to a linear transformation, which makes any attempt of interpretation futile.
It is important to make sure that a method with identifiable features is used, if the goal is to interpret them. This is another difference to Goal~\features, where identifiability may be much less important \textnew{(but see \cite{Khemakhem20NIPS,lachapelle2023synergies} who propose that identifiability may be important for learning features that transfer to new tasks).}

\paragraph{Methods}

Latent variable models provide a widely used framework. \textnew{Each latent variable or component can hopefully be interpreted as corresponding to some phenomenon or mechanism underlying the data.} For example, in neuroscience, brain activity can often be decomposed into a number of linear components \cite{Smith2013functional}. \textnew{On the other hand, if the dimension of the data is reduced so that the number of components is only two or three, the distribution of the data points can also be visualized in a scatterplot or similar.}

To facilitate interpretation, simple unsupervised models based on linear components are often preferred. 
While the classic linear factor analysis and principal component analysis suffer from lack of identifiability, Independent Component Analysis provides an identifiable alternative based on non-Gaussianity \cite {Comon94,Amari98,Hyvabookspringer}. Related methods achieve identifiability by temporal dependencies \cite{CichockiAmaribook,Hyvabookspringer} or nonnegativity \cite{Cichobooknew}.

In the nonlinear case, methods related to manifold learning, such as autoencoders and VAE are also widely used. 
However, such models typically have no identifiability guarantees, and thus the interpretation of the features is highly questionable. For example, in a typical VAE, any orthogonal transformation of the features is equally good, and thus the features are only defined up to an orthogonal transformation; actually there are even stronger indeterminacies, as discussed by \cite{Khemakhem20iVAE,Hyva23patterns}. 

Nonlinear component methods are in fact particularly difficult to identify \cite{Hyva99NN}. Recent advances show, however, that they become identifiable with suitable temporal dependecies, or related departures from independent sampling \cite{Hyva23patterns,Hyva24AISM}. 
Sometimes even such deep learning feature extraction methods may give insight to the structure of the data, if suitable methods for visualization of the features are found; some examples are shown by \cite{Zhu23,zhou2020learning}.

In addition to component models, other frameworks can be used. For example, causal discovery attempts to learn the causal connections and directions between the variables \cite{Shimizu06JMLR,Hoyer09,peters2017elements}. This is clearly interesting from the viewpoint of understanding the variables in terms of a system with interactions, and it is paramount from the viewpoint of designing interventions.  Data points can also be divided into clusters, and each cluster can be given an interpretation in terms of a subpopulation of individuals, subtypes of diseases, spatial regions, or similar.

Arguably, this goal strongly overlaps with classic statistical parameter estimation. Estimating a statistical model with a small number of parameters that have some meaning is a well-known method for understanding data. As a trivial example, simply estimating the mean and variance will give some information about the data which is easy to interpret. The discussion \textnew{just given} actually largely considered estimation of statistical models, but focusing on general-purpose models estimated on high-dimensional big data, which is the focus of machine learning.

\paragraph{Validation}

Unfortunately, validation of such results is \textnew{difficult, almost by definition.} The ``understanding'' that is the goal here should happen in the brains of humans who receive the results of the unsupervised learning as input. It is thus difficult to validate if any understanding was obtained. Asking domain experts is often the only proper way of validating these results.

In the case of causal discovery, the situation is a bit different since there is often a ground truth that one tries to recover, i.e.\ the true causal directions between the variables. In principle, that direction can often be measured by experiments, but that maybe very difficult; the difficulty of performing such experiments is the very motivation of developing causal discovery methods. Likewise, in some cases, ground truth clusters or even components may be known in some special cases.

\section{Discussion}

\subsection{Connections between the goals}

As already pointed out, some of the goals are related. An approximation of the pdf learned in Goal \pdf\ can be directly used for sampling in Goal \sampling, although sampling can also be achieved in other ways. Likewise, learning features helps both in downstream tasks in Goal \features\ and in understanding the underlying structure of the data as in Goal \understanding, although the optimal features may be different in the two cases.

Some of the goals are clearly orthogonal in the sense that one goal can be achieved without advancing towards another. In particular, Goals \pdf\ and \sampling\ need a nonlinear function approximator, but the approximator can be a total black box whose inner structure is not available. In such a case, the approximator would not enable understanding the data as in Goal \understanding, and might not even yield any features as in Goal \features.
In this sense, Goals  \pdf\ and \sampling\ are orthogonal to the Goals \features\ and \understanding. In practice, though, if the density is approximated and the sampling is enabled by a neural network, its hidden layers would usually be available as features. Still, such a neural network might not enable any understanding of the data because of the difficulties in interpreting neural networks.

There might even be some contradiction between the goals in the sense that achieving one goal makes achieving another goal more difficult. I think this would in particular be the case regarding the goal of understanding the data (Goal \understanding), which is best achieved by using simple, even linear functions, while such linear functions would not always be very useful for the other goals.


\subsection{Criticism of this list and possible further goals}

I make no claim that the list of the four goals above is exhaustive or definitive. Further goals can certainly be found, depending on what methods are actually considered to belong to unsupervised machine learning, and some of the goals here may also be found to be ill-formulated.

One objection that could be made is that these four goals are not comparable in their practical relevance. More precisely, it could be argued that density estimation (Goal~\pdf) is not a real goal of unsupervised learning, but only an intermediate goal for something else that needs to be specified. This is in contrast to Goals \sampling--\understanding, which all give outputs to the user that solve a real problem --- at least after the downstream task in Goal~\features\ is also performed. Perhaps the main justification for including Goal~\pdf\ in this list is that it is so fundamental from the viewpoint of theory and it has so many different applications, that it makes sense to consider it a separate goal.

Furthermore, causal learning has many different aspects some of which may not fit the four goals given here. While above I considered it from the viewpoint of understanding the data, it is also useful, for example, in reinforcement learning, or in general action selection, where it is important to intervene on the causes instead of the effects \cite{seitzer2021causal,dillies2025better}. It could thus be argued that causal discovery is a separate goal of unsupervised learning. 

Data compression could also be seen as a separate goal. \textnew{This would not be about simply doing PCA in order to better classify data, but performing compression, typically in some feature space, for facilitating data transmission or storage.} It takes many forms, but one specific one would be vector quantization, in the particular sense of coding $n$-dimensional real-valued vectors \textnew{corresponding to some features} as binary strings. However, this may not be a topic of great interest in the machine learning community at the moment. 
\textnew{Another candidate goal is data restoration or cleaning, including denoising, restoring missing values, and similar tasks.}

\subsection{Related work}

\textnew{The goals in the paper are closely related to the work by \cite{theis2016note}. However, they consider the problem of evaluating a given statistical generative model and show how different evaluation methods (similar to what we discussed above) lead to different rankings of such models. They don't seem to consider the idea that the learning could have completely different goals. Nevertheless, they essentially propose something like my Goals \#1, \#2, and \#3 from the viewpoint of evaluating probabilistic generative models; I add Goal \#4 (\understandingtext).}

\section{Conclusion}

As an answer to the question in the title, I proposed that unsupervised learning does not have a single goal. Instead, it has different goals which can be quite different and orthogonal to each other. I gave a list of four goals, but this list can be extended in future work. It would also be interesting to see if a more rigorous taxonomy of the goals could be developed. 

\subsection*{Conflict of Interest Statement}

The Author has no conflicts of interest to declare.


\end{document}